\documentclass{bmvc2k}
\pdfoutput=1

\usepackage[utf8]{inputenc}

\newcommand{\approachname}{LiPo-LCD}
\newcommand{\approachlongname}{Lines and Points Loop Closure Detection}

\title{\approachname: Combining Lines and Points for Appearance-based Loop Closure Detection}

\addauthor{Joan P. Company-Corcoles}{joanpep.company@uib.es}{1}
\addauthor{Emilio Garcia-Fidalgo}{emilio.garcia@uib.es}{1}
\addauthor{Alberto Ortiz}{alberto.ortiz@uib.es}{1}

\addinstitution{
 Department of Mathematics and Computer Science, University of the Balearic Islands, and \\ IDISBA (Institut d'Investigacio Sanitaria de les Illes Balears),\\
 Palma de Mallorca, Spain
}

\def\etal{\emph{et al}\bmvaOneDot}

\runninghead{Company-Corcoles \etal{}}{Combining Lines and Points for LCD}

\begin{document}

\maketitle

\begin{abstract}
Visual SLAM approaches typically depend on loop closure detection to correct the inconsistencies that may arise during the map and camera trajectory calculations, typically making use of point features for detecting and closing the existing loops. In low-textured scenarios, however, it is difficult to find enough point features and, hence, the performance of these solutions drops drastically. An alternative for human-made scenarios, due to their structural regularity, is the use of geometrical cues such as straight segments, frequently present within these environments. Under this context, in this paper we introduce LiPo-LCD, a novel appearance-based loop closure detection method that integrates lines and points. Adopting the idea of incremental Bag-of-Binary-Words schemes, we build separate BoW models for each feature, and use them to retrieve previously seen images using a late fusion strategy. Additionally, a simple but effective mechanism, based on the concept of island, groups similar images close in time to reduce the image candidate search effort. A final step validates geometrically the loop candidates by incorporating the detected lines by means of a process comprising a line feature matching stage, followed by a robust spatial verification stage, now combining both lines and points. As it is reported in the paper, LiPo-LCD compares well with several state-of-the-art solutions for a number of datasets involving different environmental conditions.

\end{abstract}

\section{Introduction}
\label{sec:intro}
Simultaneous Localization and Mapping (SLAM) is a fundamental task in autonomous mobile robotics. Regardless of the sensor used to perceive the environment, unavoidable noise sources always interfere, leading to errors in the map and the robot's pose calculations, resulting in inconsistent representations. To overcome this problem, SLAM systems usually rely on \textit{loop closure detection} (LCD) methods to recognize previously seen places. These detections provide additional constraints that can be used to correct the accumulated drift. When cameras are involved, these methods are referred to as \textit{appearance-based} loop closure detection approaches~\cite{Angeli2008b,Milford2012SeqSLAM,Cummins2011FabMap2,Galvez2012BagsOf,khan2015ibuild,Garcia2018Ibow,Bampis2018Fast}.

It is well known that many visual SLAM solutions rely on point features because of their wider applicability in general terms~\cite{Mur2017ORBSLAM2, LIM2014Realtime}. Human-made environments, however, can lack texture and thus give rise to a low number of detected features. Nevertheless, precisely because of their nature, these environments usually exhibit structural regularities that can be described using richer features such as lines, which can be more robust and less sensitive to illumination changes. Several solutions can be found in the literature describing approaches combining both kinds of features, points and lines~\cite{Pumarola2017PLSLAM,zhang2019lap}. However, despite their success, most of them rely exclusively on feature points during the LCD stage, discarding information about lines that may be useful to improve the association performance for textureless environments. Other approaches opt for using holistic image representations~\cite{Milford2012SeqSLAM,Arroyo14IROS,Sunderhauf11Brief}, which can be faster to compute but less tolerant to visual changes, while, lately, solutions based on Convolutional Neural Networks (CNNs)~\cite{snderhauf2015performance,Arroyo2016Fusion,arandjelovic2016netvlad} have shown to exhibit enhanced robustness and general performance, although they are still disengaged from real-time SLAM problems~\cite{Tsintotas2019Probabilistic,Garcia2018Ibow}. This is because they tend to require significant computational resources, e.g. on-board GPU, which makes them not suitable for mobile robotics in all cases.

The Bag of Words (BoW) model \cite{Sivic2003Video,Nister2006}, in combination with an inverted file, is arguably the most used indexing scheme for appearance-based loop closure detection~\cite{Lowry2016,GarciaFidalgo2015Survey}. Depending on how the visual vocabulary is generated, BoW-based solutions can be classified into off-line and on-line approaches. Off-line solutions generate the visual dictionary during a training phase~\cite{Cummins2011FabMap2,Galvez2012BagsOf,Mur2014Fast}, what can be high time-consuming, while the general application of the resulting vocabulary becomes highly dependent on the diversity of the training set. As an alternative, there are approaches that propose to generate the dictionary on-line~\cite{Angeli2008b,Labbe2013Appearance,khan2015ibuild,Garcia2018Ibow,Tsintots2018Assigning,Tsintotas2019Probabilistic}. Moreover, binary descriptors~\cite{Galvez2012BagsOf,Mur2014Fast,Garcia2018Ibow} have emerged recently as an alternative to real-valued descriptors for BoW models~\cite{Cummins2011FabMap2,Angeli2008b,Labbe2013Appearance}, since they offer advantages in terms of computational time and memory requirements. Additionally, similarity calculations can be performed using the Hamming distance or an of its variations, what can be efficiently implemented in modern processors.

\begin{figure}[tb]
    \begin{center}
    \includegraphics[width=1.0\linewidth]{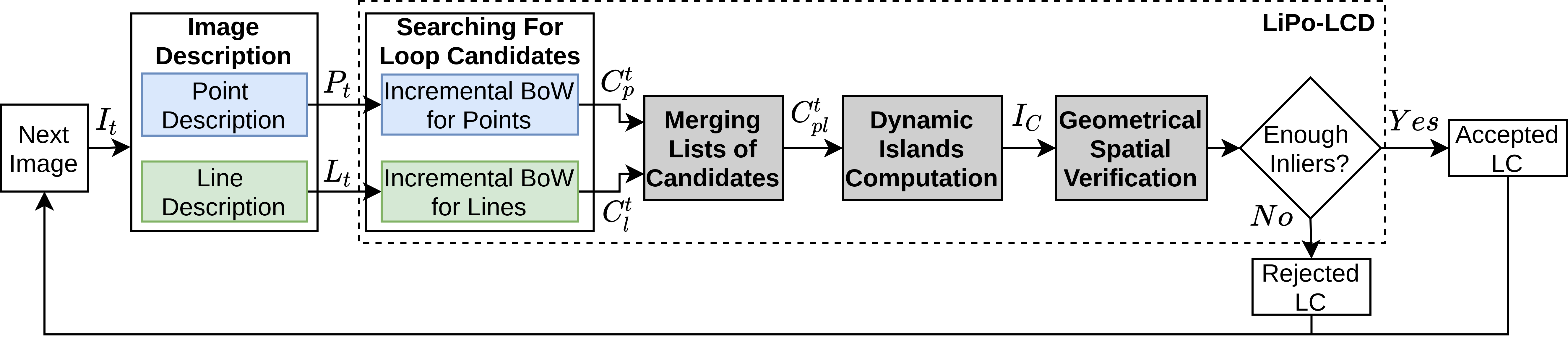}
    \end{center}
    \vspace{-2.5mm}
    \caption{General overview of the proposed loop closure detection system.}
    \label{fig:pipeline}
\end{figure}

Under this context, in this paper, we introduce \emph{\approachlongname} (\approachname), a novel appearance-based loop closure detection approach which combines points and lines. For a start, both features are described using binary descriptors. Next, an incremental BoW scheme is used for feature indexing. Lines and points are maintained separately into two incremental visual vocabularies and employed in parallel to obtain loop closure candidates efficiently. To combine the information provided by the two vocabularies, we propose a late fusion method based on a ranked voting system. Finally, to discard false positives, we improve the typical spatial verification step integrating lines into the procedure through: (1) a line matching strategy which includes structural information to achieve a higher number of matching candidates; and (2) a line representation tolerant to occlusions, which is combined with points into an epipolarity analysis step. A set of experiments validating \approachname{} and characterizing its performance against several state-of-the-art solutions is reported at the end of the paper.

Our approach follows a dual scheme to combine points and lines, such as the solutions proposed by~\cite{gomezojeda2017plslam,Zuo2017Robust}. Nonetheless, \approachname{} takes advantage of an incremental BoW strategy and incorporates lines into the spatial verification procedure that does not require map information, increasing its ability to be adapted to the operating environment, requiring only a monocular camera, and improving the performance in several datasets, as shown later.

\section{Overview of the Loop-Closure Detection Approach}
\label{sec:systemoverview}

Figure \ref{fig:pipeline} illustrates the approach proposed for loop closure detection. As can be observed, incremental visual vocabularies, along with the corresponding inverted files, are maintained independently for each visual feature. When a new image is sampled, a set of line and point binary descriptors is computed and used to (1) update the corresponding visual vocabulary and (2) obtain a list of the most similar images from each vocabulary. Next, the two lists are fused using a ranked voting procedure to obtain a final set of loop-closing candidates. To avoid adjacent images from competing with each other as loop candidates, we group images close in time using the concept of \textit{dynamic island}~\cite{Garcia2018Ibow}. Among the resulting islands, the one best corresponding with the query image is selected, while its representative image is geometrically assessed against the query to accept/reject the loop. The details about the aforementioned processes can be found next.

\subsection{Image Description}
\begin{figure}[tb]
    \centering
    \begin{tabular}{@{\hspace{0mm}}c@{\hspace{1mm}}c@{\hspace{0mm}}}
    \includegraphics[width=0.49\linewidth, clip=true, trim=0 0 0 0]{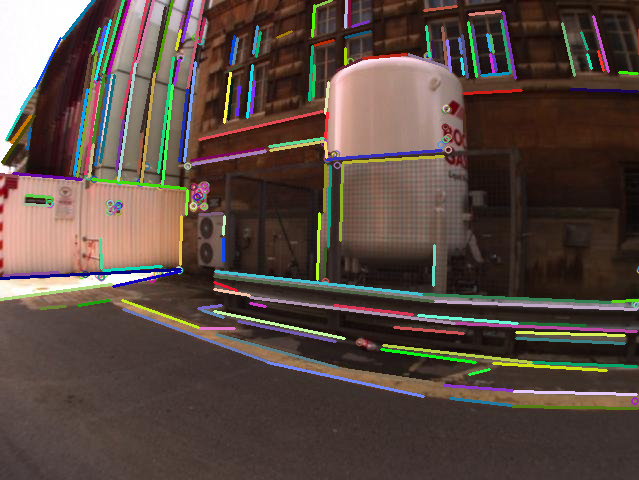} & 
    \includegraphics[width=0.49\linewidth, clip=true, trim= 0 0 0 0]{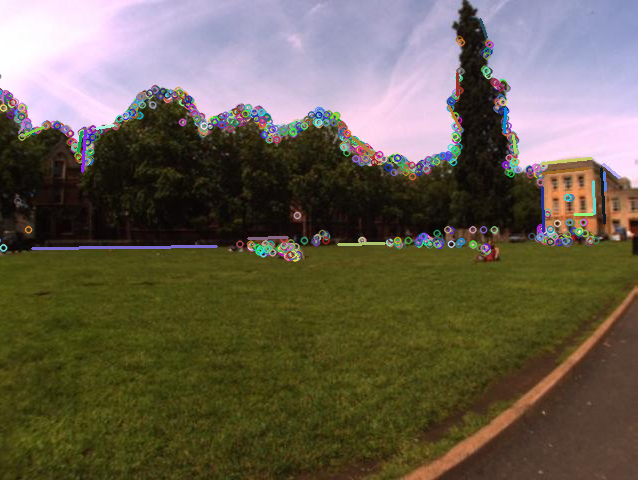} \\
    \end{tabular}
    \caption{(left) A human-made environment including a high number of lines and a low number of points. (right) An outdoor environment presenting the opposite situation.}
    \label{fig:line_and_points}
\end{figure}

As stated previously, \approachname{} describes images using lines and points. The rationale behind this approach is that the combination of multiple, complementary description techniques is a way leading to improving the performance and robustness of the loop closing method~\cite{hausler2020hierarchical}. In our solution, the image $I_t$ at time $t$ is described by $\phi(I_t)=\left\{P_{t}, L_{t}\right\}$, being $P_{t}$ a set of local keypoint descriptors and $L_{t}$ a set of line descriptors, both deriving from $I_t$. These two descriptions complement each other to make image representation more robust: while some environments may be described more distinctively using lines than points, i.e. textureless scenes, others lacking structure will benefit from keypoints, and the net result is a joint descriptor of a wider scope. Figure~\ref{fig:line_and_points} illustrates this issue for two environments.

\subsubsection{Point Description}
Given the above-mentioned advantages about binary descriptors, in this work, we have opted for detecting and describing points using ORB~\cite{rublee2011orb}. Although the proposed strategy can be used with any other binary descriptor, we employ ORB because of its robustness to rotation, scale and illumination changes~\cite{Mur2017ORBSLAM2}. The $m$ ORB descriptors found at image $I_t$ define the point descriptor as $P_t = \{p_0^t, p_1^t, \ldots, p_{m-1}^t\}$.
 
\subsubsection{Line Description}
Lines are found using the Line Segment Detector (LSD)~\cite{Grompone2010LSD}. LSD is a linear-time line segment detector that provides high-precision results and subpixel accuracy without parameter tuning. On the one hand, detected lines are described using a binary form of the Line Band Descriptor (LBD)~\cite{ZHANG2013LBD}. In the original implementation, a rectangular region centred on each line is considered. Such region is divided into a set of bands $B_i$, from which a descriptor $BD_i$ is computed contrasting $B_i$ with its neighbouring bands. On the other hand, the binary descriptor is finally obtained considering 32 possible pairs of band descriptors $BD_i$ within the support region. Each pair is compared bit by bit, generating an 8-bit string per pair. A final 256-bit descriptor is generated concatenating the resulting strings for all pairs. The set of $n$ LBD binary descriptors for image $I_t$ defines the line descriptor $L_t = \{l_0^t, l_1^t, \ldots, l_{n - 1}^t\}$.

\subsection{Searching for Loop Closure Candidates}
To index and retrieve loop closure candidates, we rely on the \textit{OBIndex2} approach~\cite{Garcia2018Ibow}, a hierarchical tree structure to manage an increasing number of binary descriptors in an efficient way. This structure can then be used as an incremental BoW scheme and combined with an inverted file for fast image retrieval. The reader is referred to~\cite{Garcia2018Ibow} for further detail.

Given that \approachname{} describes all visual features using binary descriptors, we maintain two instances of \textit{OBIndex2}, one for points and one for lines. Each instance builds an incremental visual vocabulary along with an index of images for each feature. Given an image $I_t$, a parallel search is performed on each index to retrieve the most similar images of points and lines. As a result, two lists are obtained: (1) the $m$ most similar images using points $C_p^t = \{I_{p_{0}}^t, \ldots, I_{p_{m - 1}}^t\}$ and (2) the $n$ most similar images using lines $C_l^t = \{I_{l_{0}}^t, \ldots, I_{l_{n - 1}}^t\}$. Each list is sorted by, respectively, their associated scores $s_p^t(I_{t}, I_{j}^t)$ and $s_l^t(I_{t}, I_{j}^t)$, which measure the similarity between the query image $I_t$ and the image $I_j$. Since the range of these scores varies depending on the distribution of visual words for each vocabulary, they are mapped onto the range [0,1] using \emph{min-max normalization} as follows:
\begin{equation}
\label{eq_norm_scores}
\tilde{s}_{k}^{\ t}\left(I_{t}, I_{j}^t\right) = \frac{\displaystyle s_{k}^{t}\left(I_{t}, I_{j}^t\right) - s_k^t\left(I_{t}, I_{min}^t\right)}{s_k^t\left(I_{t}, I_{max}^t\right)-s_k^t\left(I_{t}, I_{min}^t\right)}, \;\;
\end{equation}
where $s_k^t\left(I_{t}, I_{min}^t\right)$ and $s_k^t\left(I_{t}, I_{max}^t\right)$ respectively corresponds to the minimum and the maximum scores of an image candidate list, being $k \in \{p, l\}$. Images whose normalized score $\tilde{s}_k^{\ t}$ is lower than a threshold are discarded to limit the maximum number of candidates. Additionally, the current image descriptors are used to update the visual vocabularies appropriately.

\subsection{Merging Lists of Candidates}
The two resulting lists $C_p^t$ and $C_l^t$ provide loop closure candidates from each individual perspective. Thus, the next step is to combine both lists to obtain an overall overview of possible candidates but considering lines and points altogether. In this regard, the literature comprises multiple techniques to combine multimodal information for image retrieval~\cite{Bhowmik2014Efficient}. These can generally be categorized into two schemes, namely \textit{early} and \textit{late fusion}: while the former combines all features into a single representation before being processed, the latter works at the decision level, combining the outputs produced by different retrieval systems. In our proposal, given the heterogeneity of the features to combine, we rely on a late fusion approach that employs a ranked voting system based on the Borda count~\cite{Seyoon1999AnEffective} to merge lists of candidates. This is a simple data fusion form based on democratic election strategies: first, a set of voters rank a list of fixed candidates on the basis of their preferences; scores are next given to each candidate in inverse proportion to their ranking; finally, once all votes have been emitted, the candidate with the highest number of votes wins. In \approachname{}, two independent voters, one for each visual vocabulary, emit an different-size ordered list of candidates $C^t_k$. The number of candidates $c$ to vote for is set as the minimum length of the two lists. Next, top-$c$ images on each list $C^t_k$ are ranked with a score $b_k$ as:
\begin{equation}
    b_k(I^t_{i}) = (c - i)\,\tilde{s}_{k}^{\ t}\left(I_{t}, I_{i}^t\right)\,,
\label{eq:borda}
\end{equation}
where $i$ denotes the order of the image $I_i$ in the list $C^t_k$ and $\tilde{s}_{k}^{\ t}\left(I_{t}, I_{i}^t\right)$ is the normalized score of the image in that list. For each image that appears in both lists, a combined Borda score $\beta$ is computed as the geometric mean of the individual scores:
\begin{equation}
    \beta(I^t_{i}) = \sqrt{b_p(I^t_{i}) \, b_l(I^t_{i})}\,.
\label{eq:bordac}
\end{equation}

We employ the geometric mean instead of the arithmetic mean to reduce the influence of false positives in one of the lists. An integrated image list $C_{pl}^t$ results next by sorting the scores $\beta(I^t_{i})$ of all the retrieved images. This list merges information from the two visual vocabularies, independently of the number of features detected in the current environment. Finally, to deal with the fact that some environments mostly exhibit one type of feature, images that only appear in one list are also incorporated into $C_{pl}^t$, although penalized.

\subsection{Dynamic Islands Computation}
In pursuit of selecting a final loop closure candidate, in this stage we verify the temporal consistency of the images retrieved in $C_{pl}^t$. To this end, we rely on the concept of \textit{dynamic islands} used by iBoW-LCD~\cite{Garcia2018Ibow}. This method permits to avoid images competing among them as loop candidates when they come from the same area of the environment. A dynamic island $\Upsilon_{n}^{m}$ groups the images whose timestamps range from $m$ to $n$. Initially, a set of islands $\Gamma_{t}$ for the current image $I_t$ is computed considering images in the list $C_{pl}^t$ sequentially: every image $I_{i} \in C_{pl}^{t}$ is either associated to an existing island  $\Upsilon_{n}^{m}$ if the image timestamp lies in the $[m,n]$ interval or else is used to create a new island. After processing all images in $C_{pl}^{t}$, a global score $g$ is computed for each island as:
\begin{equation}
    g(\Upsilon_{n}^{m}) = \frac{\displaystyle \sum_{i=m}^n \beta(I^t_{i})}{n - m + 1}\,.
    \label{eq:ibow_island_score}
\end{equation}

Unlike~\cite{Garcia2018Ibow}, where only points are considered, in \approachname{}, score $g$ is the average of the Borda scores of the images belonging to the island, integrating both points and lines. Finally, the resulting set of islands $\Gamma_{t}$ is sorted in descending order according to $g$. Next step is to select one of the resulting islands, denoted by $\Upsilon^{*}(t)$, to determine which area of the environment is the one most likely closing a loop with $I_t$. iBoW-LCD makes use of the concept of \textit{priority islands}, defined as the islands in $\Gamma_{t}$ that overlap in time with the island selected at time $t-1$, $\Upsilon^{*}(t - 1)$. This is inspired by the fact that consecutive images should close loops with areas of the environments where previous images also closed a loop. iBoW-LCD selects, as a final island, the priority island with the highest score $g$, if any. However, this approach is just based on the appearance of the images and, therefore, due to perceptual aliasing, it might produce incorrect island associations in some human-made environments. For this reason, \approachname{} proposes a simple but effective modification of the original approach that only retains an island for the next time step if the final selected loop candidate satisfies the spatial verification procedure explained in Section~\ref{spatial_ver_sect}. Once the best island $\Upsilon^{*}(t)$ has been determined, the image $I_c$ with the highest Borda score $\beta$ of $\Upsilon^{*}(t)$ is selected as its representative and evaluated in the next verification stage.

\subsection{Spatial Verification}
\label{spatial_ver_sect}

Although the BoW scheme is a good starting point to find loop closure candidates, to finish, we perform a final geometric check to take into account the spatial arrangement of the image features and avoid perceptual aliasing. This final step comprises an epipolarity analysis between the current image $I_t$ and the loop candidate $I_c$ on the basis of the number of inliers that support the roto-translation of the camera (after computing the fundamental matrix $F$ using RANSAC). If the number of inliers is not high enough, the loop hypothesis is rejected.

The epipolarity analysis is typically carried out using a putative set of point matchings. However, as stated along this paper, point features might not be helpful because of the nature of the environment, and hence integrating lines into the geometric check can be useful, apart from the fact that straight segments can tolerate partial occlusions. To this end, \approachname{} makes use of (1) a novel line feature matching approach and (2) incorporates these line matchings, together with point matchings, into the geometric check. To match points, we make use of the available ORB descriptors, the Hamming distance and the Nearest Neighbour Distance Ratio (NNDR)~\cite{Lowe2004}.

\subsubsection{Line Feature Matching}
\label{sec:lfm}

Although NNDR is normally useful to discard false matchings between keypoints, it performs poor in respect to line descriptors matching, especially in human-made environments where line descriptors tend to be affected by perceptual aliasing~\cite{ZHANG2013LBD}. To enhance line matching performance, the authors of~\cite{ZHANG2013LBD} combine structural and appearance information in a relational graph. Despite their good results, their approach requires a high amount of memory and does not escalate well with the number of lines. In this work, we propose a much simpler but effective method to combine structural and appearance information for line feature matching. First, for each line descriptor $l_i^t$ in the current image $I_t$, we retrieve an ordered list of the most similar line descriptors of the candidate image $I_c$. Next, to deal with camera rotations, we compute a global rotation $\theta_g$ between the two frames as explained in~\cite{ZHANG2013LBD}. $\theta_g$ is next used to compute the relative orientation $\alpha_i^j$ between each pair of lines as:
\begin{equation}
    \alpha_i^j = |\theta_i^t - \theta_j^c + \theta_g|\,,
\label{eq:alpha}
\end{equation}
being $\theta_i^t$ the orientation of the line on the current image and $\theta_j^c$ the orientation of their corresponding line in the list. For each list, all line matchings with high values of $\alpha_i^j$ are discarded, and, as a result, a filtered list of line candidate matchings is obtained. To generate the final set of line matchings, we choose the two most similar surviving nearest neighbours from each list and apply the NNDR test.

\subsubsection{Epipolar Geometry Analysis Combining Points and Lines}
Works described in~\cite{Pellejero2004Automatic, Bay2005Wide} compute the fundamental matrix $F$ from homographies estimated from line segment matchings across images, provided these segments lie in at least two different planes. \approachname{} makes use of a simpler but effective approach that avoids this constraint. On the one hand, differently to other representations that can be found in the literature~\cite{Zhang2011Building, Zhou2015Struct, Li2016Joint}, in this work, line segments are represented by their endpoints. On the other hand, endpoints are first matched between matching lines and next regarded as additional point correspondences for $F$ computation. To associate segment endpoints (taking into account that a starting point of a line might correspond to the end point of the line in the other image), we select that pair that minimizes the rotation between lines using lines orientation and the global rotation $\theta_g$, as computed in Eq.~\ref{eq:alpha}. We consider a candidate line matching as an inlier if at least one endpoint pair supports the geometric model.

\section{Experimental Results}
\label{sec:experimental_results}
In this section, we evaluate the performance of \approachname{} using several public datasets. \approachname{} is also compared against some state-of-the-art solutions. All experiments were performed on an Intel Core i7-9750H (2.60 GHz) processor with 16 GB RAM.

\subsection{Methodology}

\begin{figure}[tb]
    \begin{center}
    \includegraphics[width=1.0\linewidth]{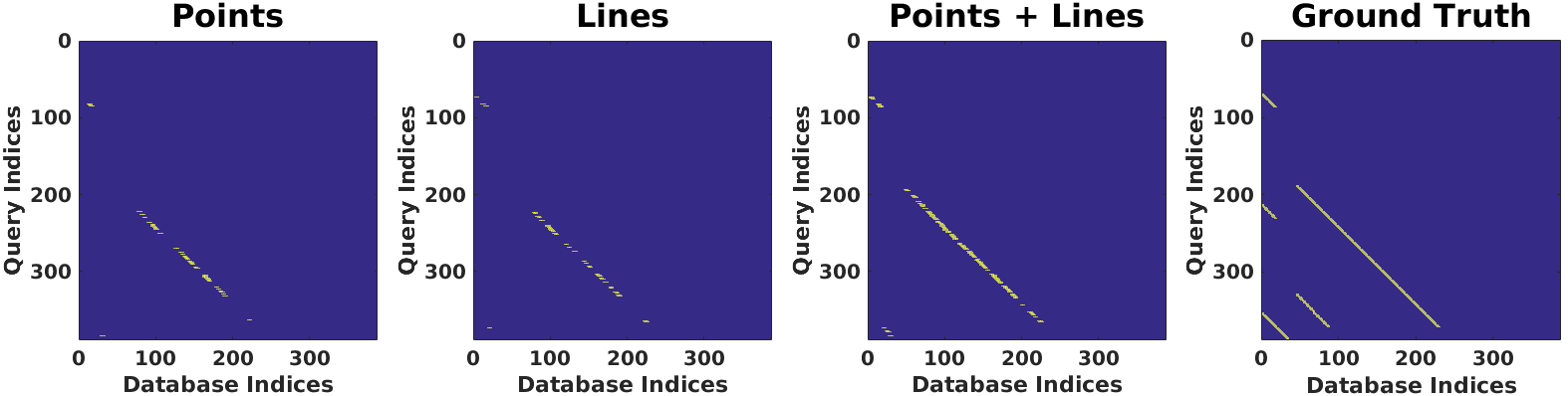}
    \end{center}
    \vspace{-2.5mm}
    \caption{Loop closure detections found in the L6I dataset using different visual features (Points, Lines, Points + Lines), and the associated ground truth. White dots represent a loop closure detected.}
    \label{fig_likelihood_mat}
\end{figure}

Precision-recall metrics are used to evaluate the system. Given that false detections can be critical if \approachname{} is used in a real SLAM solution, we are especially interested in observing the maximum recall that can be achieved at 100\% precision. OBIndex2 and iBoW-LCD were configured as explained in~\cite{Garcia2018Ibow}. The rest of the approaches shown in this section were executed using the default parameters proposed by their original authors. The following datasets were considered to validate \approachname{}: CityCentre~\cite{Cummins2008FABMAP} (CC), EuRoC Machine Hall 05~\cite{burri2016euroc} (EuR5), KITTI 00~\cite{Geiger2012} (K00), KITTI 06~\cite{Geiger2012} (K06), Lip6Indoor~\cite{Angeli2008b} (L6I), Lip6Outdoor~\cite{Angeli2008b} (L6O) and Malaga 2009 Parking 6L~\cite{blanco2009collection} (MLG). These datasets encompass a wide range of environments including, for instance, urban and indoor scenarios, which are usually rich in lines, or outdoor scenarios, where points predominate over lines. For each dataset, we use the ground truth provided by the original authors except for the KITTI sequences, where we employ the one provided by~\cite{Arroyo14IROS}, and the EuR5 and MLG datasets, where we use the files provided by~\cite{Tsintotas2019Probabilistic}.

\subsection{General Performance}
\label{sec:General_performance}

\begin{figure}[tb]
    \begin{center}
    \includegraphics[width=0.53\linewidth, clip,trim={0.0cm 0.5cm 0.0cm 0.4cm}]{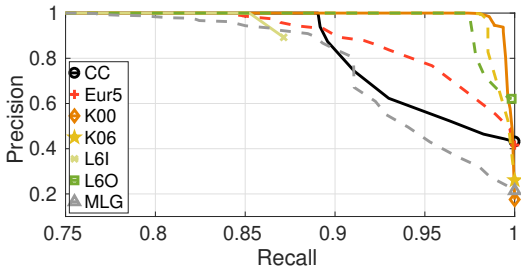}
    \end{center}
    \vspace{-3.5mm}
    \caption{P-R curves for each dataset. P is 1.0 for all R values lower than 0.75.}
    \label{fig:pr}
\end{figure}

\begin{table}[tb]
\begin{center}
\begin{tabular}{c|c|c|c|c|c|c|c|}
\cline{2-8} 
& \textbf{CC} & \textbf{EuR5} & \textbf{K00}  & \textbf{K06} & \textbf{L6I} & \textbf{L6O} & \textbf{MLG} \\ \hline \hline
 \multicolumn{1}{|l||}{\textbf{NNDR}} & 8.54 & 8.91 & 15.06 & 12.35 & 7.43 & 3.47 & 4.24\\ \hline
\multicolumn{1}{|l||}{\textbf{Proposed}}  & 18.15 & 19.21 & 25.37 & 22.51 & 10.14 & 8.45 & 11.34 \\ \hline
\end{tabular}
\end{center}
\caption{Average number of line inliers after the epipolar geometric analysis using NNDR and the proposed line feature matching method.}
\label{tab:matching_results}
\end{table}

\begin{table}[tb]
\begin{center}
\begin{tabular}{c|c|c|c|c|}
\cline{2-5}
& FE & VU & SC & SV \\ \hline \hline
\multicolumn{1}{|l||}{Points} & 18.05 & 183.16 & 146.73 & - \\
\multicolumn{1}{|l||}{Lines} & 17.60 & 23.76 & 18.13 & - \\
\multicolumn{1}{|l||}{Parallel} & 19.05 & 196.58 & 159.01 & 15.09 \\ \hline
\end{tabular}
\end{center}
\caption{Average response time (ms) per image, calculated for each part of the pipeline. These times were computed over the K00 dataset. FE: Feature Extraction; VU: Vocabulary Update; SC: Search for Candidates; SV: Spatial Verification.}
\label{tab-timesK00}
\end{table}

First, we validate the combination of points and lines proposed in this work. To this end, Fig.~\ref{fig_likelihood_mat} shows the loop closures detected by \approachname{} using points, lines and both features, as well as the ground truth for the L6I dataset, whose images are poor in feature points. As can be observed, system performance increases when points and lines are used together as visual features. To measure the global performance of the system, Fig.~\ref{fig:pr} shows precision-recall curves for \approachname{} and for each dataset. As can be observed, high recall rates are always achieved while maintaining the precision at 100\%. Moreover, \approachname{} exhibits very stable behaviour in all cases.

Next, we evaluate our novel line feature matching strategy. For that purpose, we compute the average number of line inliers on each dataset using either a classical NNDR approach for lines and our approach. Results are shown in Table~\ref{tab:matching_results}. As can be seen, the proposed line matching technique achieves a higher number of inliers in all datasets, even in sequences with severe appearance changes.
    
Finally, we evaluate the performance of \approachname{} in terms of computational times. The results obtained can be found in Table~\ref{tab-timesK00}. We show results for K00 since it is the largest dataset considered in this work. We measure the average execution time in milliseconds for each stage of the pipeline, not taking into account times for merging lists of candidates and island selection, since they are negligible. The average response time of the whole system per image turns out to be 389.79 ms using a parallel implementation. As can be observed, feature extraction steps are very fast in all cases. The vocabulary update and the search for candidates steps are slower for points, due to the number of features to handle on each case. The spatial verification stage is always performed using points and lines together, and, hence, times for each feature separately are not available.

\subsection{Comparison with Other Solutions}

\begin{table}[tb]
\begin{center}
\resizebox{0.88\textwidth}{!}{%
\begin{tabular}{c|c|c|c|c|c|c|c|}
\cline{2-8} 
& \textbf{CC} & \textbf{EuR5} & \textbf{K00}  & \textbf{K06} & \textbf{L6I} & \textbf{L6O} & \textbf{MLG} \\ \hline \hline
 \multicolumn{1}{|l||}{\textbf{Bampis}~\cite{Bampis2018Fast} }         & 71.14 & n.a. & 96.53 & n.a. & 52.22 & 58.32 & 87.56  \\ \hline
\multicolumn{1}{|l||}{\textbf{Gálvez-López}~\cite{Galvez2012BagsOf} }  & 31.61 & n.a. & n.a. & n.a. & n.a. & n.a. & 74.75  \\ \hline
\multicolumn{1}{|l||}{\textbf{Mur-Artal}~\cite{Mur2014Fast} }  
& 43.03 & n.a. & n.a. & n.a. & n.a. & n.a. & 81.51 \\ \hline
\multicolumn{1}{|l||}{\textbf{Cummins}~\cite{Cummins2011FabMap2} } & 38.77 & n.a. & 49.2 & 55.34 & n.a. & n.a. & 68.52 \\ \hline
\multicolumn{1}{|l||}{\textbf{Stumm}~\cite{Stumm2016Building} } & 38.00 & n.a. & n.a. & n.a. & n.a. & n.a. & n.a. \\ \hline
\multicolumn{1}{|l||}{\textbf{Gomez-Ojeda}~\cite{gomezojeda2017plslam}}
& n.a. & 1.61 & 75.93 & 56.94 & n.a. & n.a. & n.a. \\ \hline \hline \hline
\multicolumn{1}{|l||}{\textbf{Tsintotas}~\cite{Tsintotas2019Probabilistic} }
& n.a. & \textbf{83.7} & 97.5 & n.a. & n.a. & 50.0 & 85.0 \\ \hline
\multicolumn{1}{|l||}{\textbf{Tsintotas}~\cite{Tsintots2018Assigning} } 
& n.a. & 69.2 & 93.2 & n.a. & n.a. & n.a. & \textbf{87.9} \\ \hline
\multicolumn{1}{|l||}{\textbf{Angeli}~\cite{Angeli2008b} }
& n.a. & n.a. & n.a. & n.a. & 36.86 & 23.59 & n.a. \\ \hline
\multicolumn{1}{|l||}{ \textbf{Zhang}~\cite{zhang2016learning} }
& 41.2 & n.a. & n.a. & n.a. & n.a. & n.a. & 82.6 \\ \hline
\multicolumn{1}{|l||}{\textbf{Gehrig}~\cite{gehrig2017visual}  }
& n.a. & 71.0 & 93.1 & n.a. & n.a. & n.a. & n.a. \\ \hline
\multicolumn{1}{|l||}{\textbf{Khan}~\cite{khan2015ibuild} }
& 38.92 & n.a. & n.a. & n.a. & 41.74 & 25.58 & 78.13 \\ \hline
\multicolumn{1}{|l||}{\textbf{Garcia-Fidalgo}~\cite{Garcia2018Ibow}}
& 88.25 & n.a. & 76.50 & 95.53 & 83.18 & 85.24 & n.a. \\ \hline \hline \hline
\multicolumn{1}{|l||}{\textbf{\approachname}}
& \textbf{89.30} & 81.94 & \textbf{97.80} & \textbf{97.38} & \textbf{ 85.24} & \textbf{ 97.31  } & 75.73 \\ \hline
\end{tabular}
}
\end{center}
\caption{Maximum recall at 100\% precision for several off-line approaches (top), on-line approaches (middle) and the proposed solution (bottom). Winners are indicated in bold face.}
\label{tab:PR_metrics}
\vspace{-3.5mm}
\end{table}

In this last section, \approachname{} is compared with other solutions. Table~\ref{tab:PR_metrics} shows the maximum recall achieved at 100\% precision for all approaches. The results reported come from the original works, except for~\cite{gomezojeda2017plslam}, which was executed by ourselves using the vocabularies and the default parameters provided by their authors. Results not available are indicated by \textit{n.a.} As can be observed, \approachname{} achieves, in most cases, a higher recall than the other solutions. This is particularly interesting regarding the L6I dataset, where the combination of points and lines allows us to increase the performance in a low-textured scenario. It is also worth mentioning that \approachname{} outperforms~\cite{gomezojeda2017plslam}, which is perhaps the most similar solution to ours.

\section{Conclusions}
\label{sec:conclusions}

In this work, we have described \approachname{}, an appearance-based loop closure detection method that combines points and lines. This combination allows us to detect loops in environments poor of feature points. Moreover, points and lines are described using binary descriptors for execution time reduction. To obtain loop closure candidates from both visual clues, we rely on a dual incremental BoW scheme. A late fusion method for merging both lists of candidates, based on the Borda count, is also proposed. The loop candidate hypothesis is finally validated by means of a geometrical check, which involves both points and lines. \approachname{} compares favourably with several state-of-the-art methods under different environmental conditions.

\section*{Acknowledgements}
This work is partially supported by EU-H2020 projects BUGWRIGHT2 (GA 871260) and ROBINS (GA 779776), and by projects PGC2018-095709-B-C21 (MCIU/AEI/FEDER, UE), and PROCOE/4/2017 (Govern Balear, 50\% P.O. FEDER 2014-2020 Illes Balears). This publication reflects only the authors views and the European Union is not liable for any use that may be made of the information contained therein.

\bibliographystyle{bmvc2k}
\bibliography{ms}
\end{document}